%% file: main_arxiv.tex
\documentclass{bmvc2k}

\usepackage{graphicx}
\usepackage{booktabs}
\usepackage{multirow}
\usepackage{floatrow}
\usepackage[capitalize]{cleveref}
\usepackage{xspace}
\usepackage{amsmath,bbm}
\usepackage{pifont}

\title{CLIP with Quality Captions: A Strong Pretraining for Vision Tasks}

\addauthor{Pavan Kumar Anasosalu Vasu}{panasosaluvasu@apple.com}{1}
\addauthor{Hadi Pouransari}{mpouransari@apple.com}{1}
\addauthor{Fartash Faghri}{fartash@apple.com}{1}
\addauthor{Oncel Tuzel}{otuzel@apple.com}{1}

\addinstitution{
 Apple
}

\runninghead{P.K.A. Vasu et al.}{Effectiveness of CLIP pretraining}

\def\eg{{\emph{e.g}\bmvaOneDot} }

\crefname{section}{Sec.}{Secs.}
\Crefname{section}{Section}{Sections}
\Crefname{table}{Table}{Tables}
\crefname{table}{Tab.}{Tabs.}

\def\datacomp{{DataComp}}
\def\datacompdr{{DataCompDR}}
\definecolor{darkgreen}{RGB}{34, 139, 34}

\input{defs}

\begin{document}

\maketitle

\begin{abstract}
  CLIP models perform remarkably well on zero-shot classification and retrieval tasks. But recent studies have shown that learnt representations in CLIP are not well suited for dense prediction tasks like object detection, semantic segmentation or depth estimation. 
  More recently, multi-stage training methods for CLIP models was introduced to mitigate the weak performance of CLIP on downstream tasks.
  In this work, we find that simply improving the quality of captions in image-text datasets improves the quality of CLIP's visual representations, resulting in significant improvement on downstream dense prediction vision tasks. In fact, we find that CLIP pretraining with good quality captions can surpass recent supervised, self-supervised and weakly supervised pretraining methods.
  We show that when CLIP model with ViT-B/16 as image encoder is trained on well aligned image-text pairs it obtains 12.1\% higher mIoU and 11.5\% lower RMSE on semantic segmentation and depth estimation tasks over recent state-of-the-art Masked Image Modeling (MIM) pretraining methods like Masked Autoencoder (MAE). We find that mobile architectures also benefit significantly from CLIP pretraining. A recent mobile vision architecture, MCi2, with CLIP pretraining obtains similar performance as Swin-L, pretrained on ImageNet-22k for semantic segmentation task while being 6.1$\times$ smaller. Moreover, we show that improving caption quality results in $10\times$ data efficiency when finetuning for dense prediction tasks.
\end{abstract}

%-------------------------------------------------------------------------
\section{Introduction}
\label{sec:intro}

Pretraining on a large corpus and then finetuning on the target task is a common paradigm in computer vision. In the past decade, a common pretraining strategy was to preform supervised pretraining on ImageNet. Recently, models like contrastive language-image pretraining (CLIP)~\cite{clip},  BEiT~\cite{beit}, and DINO~\cite{dino} trained on large-scale datasets have shown to learn generic and highly transferable visual representations. These pretrained models are then used as initialization and finetuned for various downstream tasks like object detection, semantic segmentation, and depth estimation.

While there exists a wide range of pretraining methods, no single method works best for all downstream tasks. Previously, CLIP pretraining has shown unsatisfactory performance for dense prediction tasks~\cite{clipft} such as object detection and depth estimation compared with MIM pretraining methods (\eg MAE~\cite{mae}) and self-supervised learning (SSL) methods (\eg DINO~\cite{dino}). 
In this work, we empirically study the following research question: \textit{Is CLIP fundamentally a poor pretraining choice compared to MIM for dense prediction tasks?}  

In CLIP pretraining, we learn an image and a text encoder to align embeddings from (image, text) pairs. The (image, text) data, collected on a large-scale through web crawling, is often noisy: the text and image content may not align well. \citet{cherti2022reproducible} have shown that the scale of the pretraining dataset significantly impacts the quality of representations learned by CLIP image and text encoders. In this work, we show that the quality of captions (text modality) significantly affects the quality of visual representations (image modality). We find that CLIP image encoder features trained with aligned captions not only transfer well for semantic discriminative tasks but also result in significant performance gains for dense prediction vision tasks.

\begin{figure}[t!]
    \centering
    \includegraphics[width=0.99\linewidth]{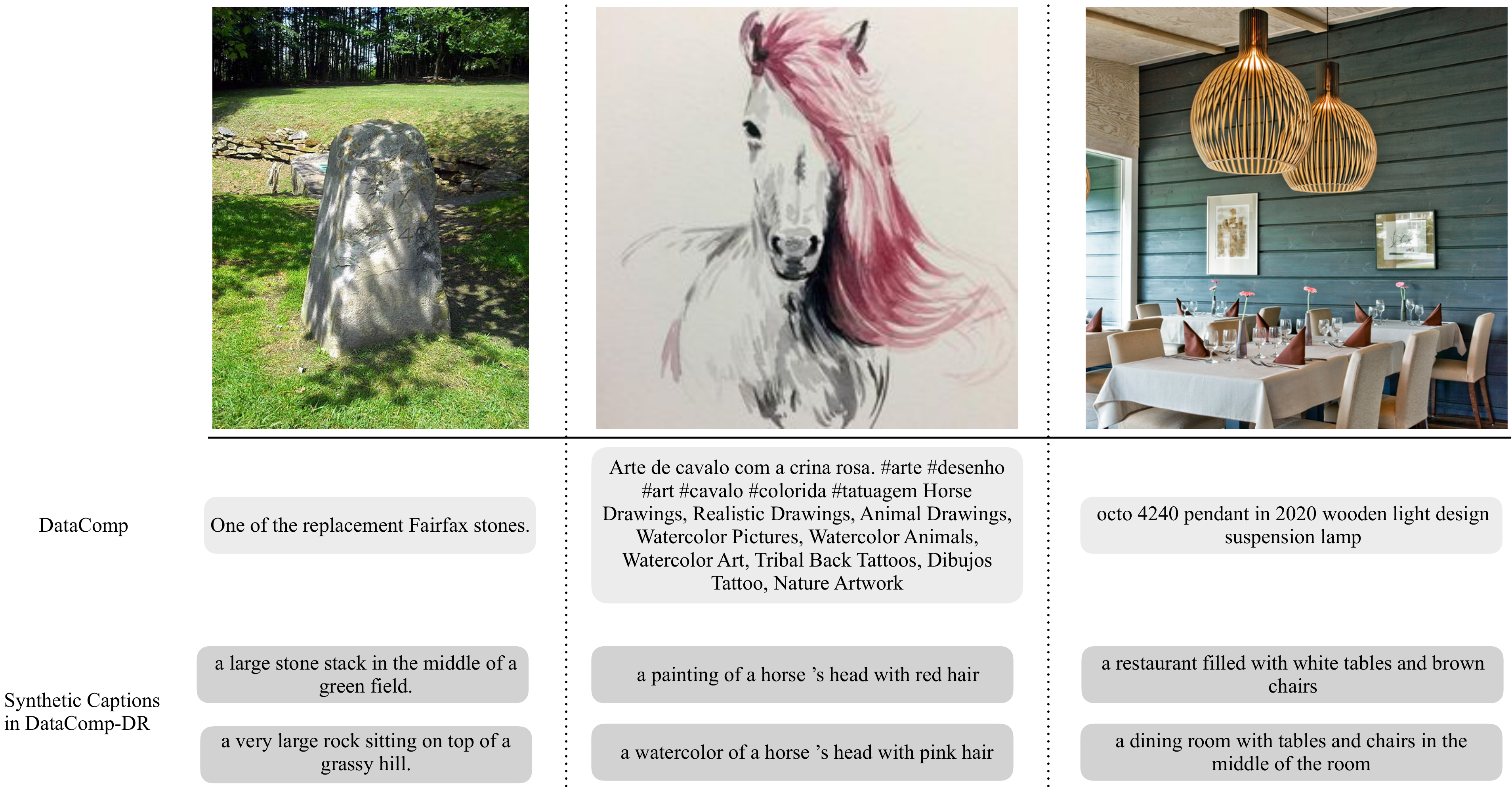}
    \caption{Qualitative examples of captions in \datacomp{} and \datacompdr{}  datasets.}
    \label{fig:caption_examples}
\end{figure}

% All results on ViT-B/16, Dataset & Caption ablation
\begin{table}[t!]
    \centering
    \resizebox{0.8\linewidth}{!}{
        \begin{tabular}{l@{\hspace*{4mm}}c@{\hspace*{4mm}}c@{\hspace*{4mm}}c@{\hspace*{4mm}}c@{\hspace*{4mm}}c@{\hspace*{4mm}}c@{\hspace*{4mm}}c@{\hspace*{4mm}}c@{\hspace*{4mm}}c}
            \toprule[1.5pt]
            \multirow{2}{*}{\textbf{Dataset}} &
            \multirow{2}{*}{\textbf{Size}} &
            \multirow{2}{*}{\parbox{2cm}{\centering \textbf{Caption Quality}}} &
            \multicolumn{2}{c}{\textbf{IN-1K (\%)}} &
            \multicolumn{2}{c}{\textbf{Detection (mAP)}} &
            \multicolumn{2}{c}{\textbf{Segmentation}} &
            \textbf{Depth}
            \\
            \cmidrule(lr){4-10}
            & & & Linear & F.T. & BBox & Mask &
            mIoU & mAcc & RMSE($\downarrow$)
            \\
             \midrule[1.25pt]
        OpenAI-WIT & 400M &  & 79.5 & 82.9 & 45.0 & 39.8 & 49.5 & - & 0.416 \\ 
        ALIGN$^\dagger$ & 1.1B & + & 80.4 & 85.4 & 47.5 & 42.3 & 51.2 & 62.8 & 0.380 \\
        \datacomp{} & 1.28B & ++ & 81.8 & 85.9 & 48.4 & 43.4 & 52.5 & 64.2 & 0.358 \\
        \datacompdr{}  & 1.28B & +++ & \textbf{82.4} & \textbf{86.4} & \textbf{48.9} & \textbf{43.6} & \textbf{53.9} & \textbf{65.2} & \textbf{0.339} \\

            \bottomrule[1.5pt]
        \end{tabular}
    }
    \caption{Results on downstream dense prediction tasks for CLIP models trained on different image-text datasets. $^\dagger$ refers to our reproduction of ALIGN dataset. All results are for a ViT-B/16 model, pretrained on the respective datasets at 224x224 resolution. All the models are trained for the same duration. For detection and segmentation, we use Mask-RCNN and UperNet heads respectively. For depth prediction, we follow~\cite{xie2023darkmim}.}
    \label{tab:dataset_abl}
\end{table}

To analyze the effect of caption quality on CLIP models, we compare CLIP models trained on four datasets with varying caption quality as shown in \Cref{tab:dataset_abl}. Since OpenAI's WIT dataset is not publicly available, we do not know the quality of the underlying captions. ALIGN~\cite{align} dataset comprises noisy alt-text captions. \datacomp{}~\cite{datacomp} introduced an image-text dataset where the captions have been filtered based on the CLIP score and other heuristics that improve the quality of captions. \datacompdr{}~\cite{mobileclip2024} augments the original image-text from \datacomp{} with synthetic captions generated by a CoCa~\cite{COCA} model. As shown in~\cref{fig:caption_examples}, synthetic captions are better aligned with the image than the original caption scraped from the web in \datacomp{}.
From~\cref{tab:dataset_abl}, we notice that CLIP models trained on datasets with better captions transfer significantly better to a wide range of downstream tasks.

We also find CLIP pretraining to work well for mobile architectures. Most mobile architectures such as \cite{fastvit, efficientformer, mobileone, Mobilenet_v3} are either CNN-transformer hybrid or purely convolutional in design. Pretraining methods like MAE~\cite{mae} are tailored to transformers and cannot be directly applied to mobile architectures. 
CLIP pretraining on \datacompdr{} improves the performance of mobile architectures on dense prediction tasks like object detection and semantic segmentation as discussed in~\Cref{sec:impact}. We will release our code for finetuning along with all model checkpoints.

The main contributions of our work are as follows,
% \vspace{-2pt}
\begin{itemize}
\vspace{-5pt}
\item Through systematic experimentation, we demonstrate that CLIP can learn visual features suitable for dense prediction tasks, provided that image-text pairs in the pretraining dataset are well aligned.
\vspace{-5pt}
\item We present a detailed comparison between CLIP and MAE pretraining strategies. Across various training schedules, we show CLIP pretraining can outperform MAE, provided the dataset contains good quality captions. 
\vspace{-5pt}
\item We present a detailed analysis of data scaling trends of CLIP pretraining for dense prediction tasks. We show that improving captions results in better data efficiency.
\vspace{-5pt}
\item We demonstrate that CLIP pretraining benefits even mobile architectures,
achieving state-of-the-art accuracy-latency trade-offs for vision tasks such as detection, segmentation, and depth estimation.
 
\end{itemize}

\vspace{-10pt}
\section{Background}
Learning transferable representations in computer vision is an active area of research. A common paradigm in computer vision is pretraining followed by finetuning on downstream tasks such as detection~\cite{fasterrcnn, maskrcnn, cascadercnn}, segmentation~\cite{fcn_seg, kirillov2019panoptic, upernet}, depth estimation~\cite{dpt_depth}, etc.. Supervised pretraining has been widely used where models are trained on large-scale labeled datasets~\cite{scalevit}. But acquiring accurate labels at scale is a challenge and most of the large-scale labeled datasets like JFT~\cite{scalevit} are private. 

Recently there has been significant progress in self-supervised pretraining. Most of these methods do not require accurately labeled datasets and use pretraining strategies, \eg instance contrastive learning ~\cite{chen2020big, chen2020simple, he2019moco}, puzzle solving~\cite{noroozi2017unsupervised, zhai2022position}, joint embedding~\cite{ibot, dBOT, caron2021emerging, grill2020bootstrap} or masked region reconstruction~\cite{mae, beit}. In particular, MAE~\cite{mae} has shown to learn highly transferable representations. But MAE has limitations--\citet{singh2023effectiveness} showed that smaller image encoders like ViT-B/16 do not benefit from dataset scaling. 
In~\cite{singh2023effectiveness}, MAE is followed by a second stage of weakly supervised pretraining to obtain further significant improvements on downstream tasks.

\vspace{-5pt}
\subsection{Contrastive Language-Image Pretraining (CLIP)}
CLIP~\citep{clip} is an image-text model that maps image and text
to a join embedding space.
CLIP consists of image and text encoders trained on a large-scale dataset of
paired image-text samples such that similar images and texts are mapped
close to each other and farther from dissimilar
samples~\citep{karpathy2015deep,kiros2014unifying,faghri2018vse++}.
Given a batch of $b$ image-text pairs, we denote the image and text $d$-dimensional embeddings of a CLIP model
by $\FeatImg, \FeatTxt \in \mathcal{R}^{b \times  d}$.
Let $\Similarity_\tau(\vv, \UU)=\text{Softmax}(\UU\vv/\tau) \in \mathcal{R}^{b}$ denote the similarity between an embedding vector $\vv$ and a matrix of embeddings $\UU$ normalized by Softmax with a temperature parameter $\tau$.
The CLIP loss consists of an image-to-text loss and a text-to-image loss
with the image-to-text component defined as,
\begin{align}\label{eq:contrastive}
    \LCEIT(\mathcal{B}) &= \frac{1}{b}
    \sum_{i=1}^b
    \text{CE}(
    \Similarity_{\tau}(\FeatImgi, \FeatTxt),
    \ee_i)
\end{align}
where CE denotes the Cross-Entropy loss with a $b$-dimensional one-hot vector $\ee_i$ with a 1 in $i$-th coordinate and 0's elsewhere.
Similarly, the text-to-image loss, $\LCETI$, is defined by exchanging the text and image embedding.

\vspace{-5pt}
\subsection{Importance of Data Quality for CLIP}
CLIP demonstrated significant zero-shot capabilities in downstream classification tasks without using any fine-tuning on task-specific data.
In CLIP~\cite{clip}, a multi-modal model was trained with images and their noisy text annotations from the web. Recently, datasets have scaled up significantly and CLIP pretraining has shown significant improvements in zero-shot image classification and retrieval performance~\cite{cherti2022reproducible}.

\citet{fang2022data} showed that the quality and diversity of the pretraining
data distribution of CLIP can explain its emergent zero-shot capabilities.
Building on this observation, \citet{datacomp} proposed the DataComp
benchmark for
finding the optimal CLIP training set and proposed \datacomp{}-1B
(BestPool filtering)
that substantial improved CLIP training using public datasets
compared with prior dataset LAION-2B~\citep{schuhmann2022laion}.
Recently, \citet{mobileclip2024} introduced \datacompdr{} that along with other information added synthetic captions to \datacomp{}-1B at scale which were cleaner and well-aligned.

The impact of scale and quality on downstream dense prediction tasks has not been previously studied for CLIP pretraining. \citet{nguyen2023improving} investigated improving captions to improve image retrieval and image captioning performance. \citet{clipft} observed that CLIP pretraining underperforms other pretraining methods on dense prediction tasks and suggested an additional fine-tuning step to improve CLIP image encoder. The results in \cite{clipft} are limited to early CLIP models trained on a relatively small dataset. In contrast, we find that CLIP pretraining, on the larger \datacomp{}~\cite{datacomp} is highly competitive to other pretraining works. In this work, we show that CLIP models trained on datasets with improved caption quality (\eg \datacompdr{}~\cite{mobileclip2024}) significantly improve the quality of the visual representations for downstream dense prediction tasks.

\vspace{-8pt}
\section{Analysis}
% \vspace{-5pt}
\subsection{CLIP with Synthetic Captions}
While \datacomp{}~\cite{datacomp} used filtering heuristics to obtain good quality captions, recently methods have used large vision foundation models and language models to generate high quality synthetic captions. These captions tend to be better aligned with the respective images and higher quality when compared to noisy text from the web. 
We compare finetuning performance of CLIP models pretrained on ground-truth captions and synthetic higher quality captions in recent MobileCLIP~\cite{mobileclip2024}, LaCLIP~\cite{fan2023laclip} and VeCLIP~\cite{lai2024veclip} works. LaCLIP, mostly employs LLMs to rewrite captions, VeCLIP employs LLaVA~\cite{liu2023llava} model and an LLM to caption their dataset called VeCap. MobileCLIP used a CoCa~\cite{COCA} model to caption \datacomp{}~\cite{datacomp} dataset resulting in \datacompdr{}. VeCap and \datacompdr{} contains visually enriched captions through the use of vision-language foundation models as opposed to LaCLIP that only relies on LLMs to rephrase existing captions. 
From~\Cref{tab:dataset_abl}, we see that improved caption quality results in image encoders that consistently transfer better to dense prediction tasks. We also find that CLIP models pretrained on visually enriched captions transfer better to dense prediction tasks.

\begin{table}[t!]
    \centering
    \resizebox{0.8\linewidth}{!}{
        \begin{tabular}{l@{\hspace*{4mm}}c@{\hspace*{4mm}}c@{\hspace*{4mm}}c@{\hspace*{4mm}}c@{\hspace*{4mm}}c@{\hspace*{4mm}}c@{\hspace*{4mm}}c@{\hspace*{4mm}}c}
            \toprule[1.5pt]
            \textbf{Dataset} &
            \textbf{Caption} &
            \textbf{Visual} &
            \textbf{LLM} &
            \textbf{Captions} &
            \multicolumn{2}{c}{\textbf{Detection (mAP)}} &
            \multicolumn{2}{c}{\textbf{Segmentation}}
            \\
            \cmidrule(lr){6-9}
            \textbf{Source} & \textbf{Source} & \textbf{Enrich.} & \textbf{Rewrite} & \textbf{Used} & BBox & Mask &
            mIoU & mAcc
            \\
             \midrule[1.25pt]
        \multirow{2}{*}{\datacomp{}-12M} & Ground Truth & - & - & G & 42.4 & 38.3 & 43.9 & 54.7 \\ 
                                    & CoCa~\cite{COCA}     & \cmark & \xmark & G + S & \textbf{44.6} & \textbf{39.9} & \textbf{47.8} & \textbf{59.0} \\

            \midrule[0.5pt] 
        \multirow{3}{*}{CC-12M} & Ground Truth & - & - & G & 43.8 & 39.3 & 45.5 & 55.6 \\
             & ChatGPT, Llama-7B~\cite{touvron2023llama}, & \multirow{2}{*}{\xmark} & \multirow{2}{*}{\cmark} & \multirow{2}{*}{S} & \multirow{2}{*}{\textbf{43.9}} & \multirow{2}{*}{\textbf{39.5}} & \multirow{2}{*}{\textbf{46.1}} & \multirow{2}{*}{\textbf{56.2}} \\
             & Bard, Human Annotators & & & & & & & \\

            \midrule[0.5pt]
        \multirow{2}{*}{WIT-12M} & Ground Truth & - & - & G & 41.4 & 37.4 & 42.5 & 52.4 \\
             & LLaVa~\cite{liu2023llava} + Vicuna-1.1~\cite{zheng2023judging} & \cmark & \cmark & G + S & \textbf{42.7} & \textbf{38.5} & \textbf{45.6} & \textbf{55.7} \\
            \bottomrule[1.5pt]
        \end{tabular}
    }
    \caption{Dataset ablation for CLIP models. All the datasets contain 12M samples and models are trained for the similar duration, i.e. $\sim$0.4B seen samples. All results are for a ViT-B/16 model, pretrained on the respective datasets at 224x224 resolution. For detection and segmentation, we use Mask-RCNN and UperNet heads respectively. ``G" stands for supervision on ground truth captions, ``S" stands for supervision on synthetic cations.}
    \label{tab:dataset_abl}
\end{table}

\subsection{CLIP versus MAE}
To have a fair comparison between the two popular methods, we train a CLIP model on \datacomp{}~\cite{datacomp} and \datacompdr{}~\cite{mobileclip2024} datasets and match the number of seen samples (i.e global batch size $\times$ total number of iterations) between the two pretraining methods.
In \cref{tab:clip_vs_mae_seen_samples}, for 2.05B seen samples we observe that MAE outperforms CLIP by 0.9 mAP on object detection task, but CLIP pretraining on \datacompdr{} outperforms MAE by 3.6 mIOU on segmentation task and obtains 4.4\% lower RMSE on depth estimation task. When MAE pretraining is scaled to the IG-3B dataset which contains 3B unique images (larger than \datacompdr{}) and 28K classes, we do not see any improvements in detection and segmentation tasks. Whereas CLIP pretraining overtakes MAE on object detection and obtains 4.4 further improvement in mIoU on the segmentation task. At this scale MAE outperforms CLIP pretraining only on depth estimation. When CLIP model is trained longer on \datacompdr{} at similar scale as MAWS~\cite{singh2023effectiveness}, it outperforms both MAE and MAWS on all the downstream dense prediction tasks.

\begin{table}[t!]
    \centering
    \resizebox{0.75\linewidth}{!}{
        \begin{tabular}{l@{\hspace*{4mm}}c@{\hspace*{4mm}}c@{\hspace*{4mm}}c@{\hspace*{4mm}}c@{\hspace*{4mm}}c@{\hspace*{4mm}}c@{\hspace*{4mm}}c}
            \toprule[1.5pt]
            \multirow{2}{*}{\textbf{Method}} &
            \textbf{Pre-Training} &
            \textbf{\# Seen} &
            \multicolumn{2}{c}{\textbf{Detection (mAP)}} &
            \multicolumn{2}{c}{\textbf{Segmentation}} &
            \textbf{Depth}
            \\
            \cmidrule(lr){4-8}
            & \textbf{Dataset} & \textbf{Samples} & BBox & Mask &
            mIoU & mAcc & RMSE($\downarrow$)
            \\
             \midrule[1.25pt]
        MAE~\cite{mae} & IN-1K & \multirow{3}{*}{2.05 B} & \textbf{54.0} & \textbf{46.7} & 48.1 & 58.9 & 0.383 \\ 
        \multirow{2}{*}{CLIP} & \datacomp{} &  & 52.1 & 44.9 & 49.1 & 60.6 & 0.398 \\
             & \datacompdr{} & & 53.1 & 45.7 & \textbf{51.7} & \textbf{63.3} & \textbf{0.366} \\

            \midrule[0.5pt]
        MAE~\cite{singh2023effectiveness} & IG-3B & \multirow{3}{*}{5 B} & 53.8 & 46.5 & 47.2 & 58.0 & \textbf{0.348} \\ 
        \multirow{2}{*}{CLIP} & \datacomp{} &  & 53.3 & 45.9 & 50.7 & 62.3 & 0.374 \\
             & \datacompdr{} & & \textbf{54.1} & \textbf{46.7} & \textbf{52.5} & \textbf{64.2} & 0.351 \\

            \midrule[0.5pt]
            MAWS (MAE+WSP)~\cite{singh2023effectiveness} & IG-3B       & 10 B & 53.9 & 46.6 & 50.4 & 61.5 & 0.371 \\
            \multirow{2}{*}{CLIP} & \datacomp{} & \multirow{2}{*}{12.8 B} & 53.9 & 46.4 & 52.5 & 64.2 & 0.358 \\
                       & \datacompdr{} &  & \textbf{54.6} & \textbf{47.1} & \textbf{53.9} & \textbf{65.2} & \textbf{0.339} \\ 
            \bottomrule[1.5pt]
        \end{tabular}
    }
    \caption{Results on downstream dense prediction tasks for MAE and CLIP models, when trained for the similar duration, i.e. \# of seen samples. All results are for a ViT-B/16 model, pretrained on the respective datasets at 224x224 resolution. For detection and segmentation, we use Cascade Mask-RCNN and UperNet heads respectively. For depth prediction, we follow~\cite{xie2023darkmim}.}
    \label{tab:clip_vs_mae_seen_samples}
\end{table}

\subsection{Data Scaling}
\begin{figure}[t!]
    \centering
    \includegraphics[width=0.99\linewidth]{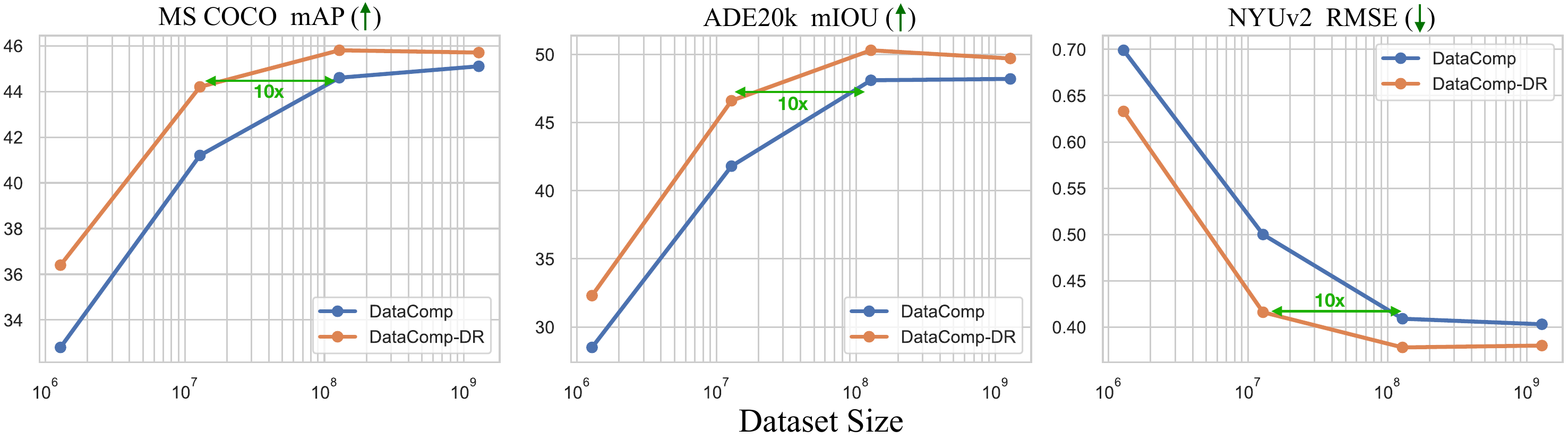}
    \caption{Data scaling trends for CLIP pretraining on \datacomp{} and \datacompdr{}. All results are for a ViT-B/16 model. Improved caption quality results in better data efficiency for learning transferable representations.}
    \label{fig:data_scaling_perf}
\end{figure}

To understand the effect of scaling dataset size we train CLIP models on \datacomp{} and \datacompdr{} on subsets ranging from 1.28M to all of 1.28B samples and compare their finetuning performance in~\Cref{fig:data_scaling_perf}. For all experiments, we pretrain for 20k iterations with global batch size of 65k (equivalent to one epoch training on 1.28B). The image encoders are fine-tuned on downstream tasks using the settings described in~\Cref{sec:experiments}. From~\Cref{fig:data_scaling_perf}, we find that improved caption quality results in better data efficiency when finetuning for dense prediction tasks. On MS COCO, CLIP model pretrained on 12.8M subset of \datacompdr{} obtains an mAP of 44.2 which is slightly (0.9\%) worse than 44.6, mAP obtained by pretraining on 128M subset of \datacomp{} for object detection task. CLIP models can be pretrained on 10$\times$ smaller subset of \datacompdr{} to obtain similar performance as pretraining on a larger subset of \datacomp{}.

\subsection{Representation Analysis}

\begin{figure}[t!]
    \centering
    \includegraphics[width=0.99\linewidth]{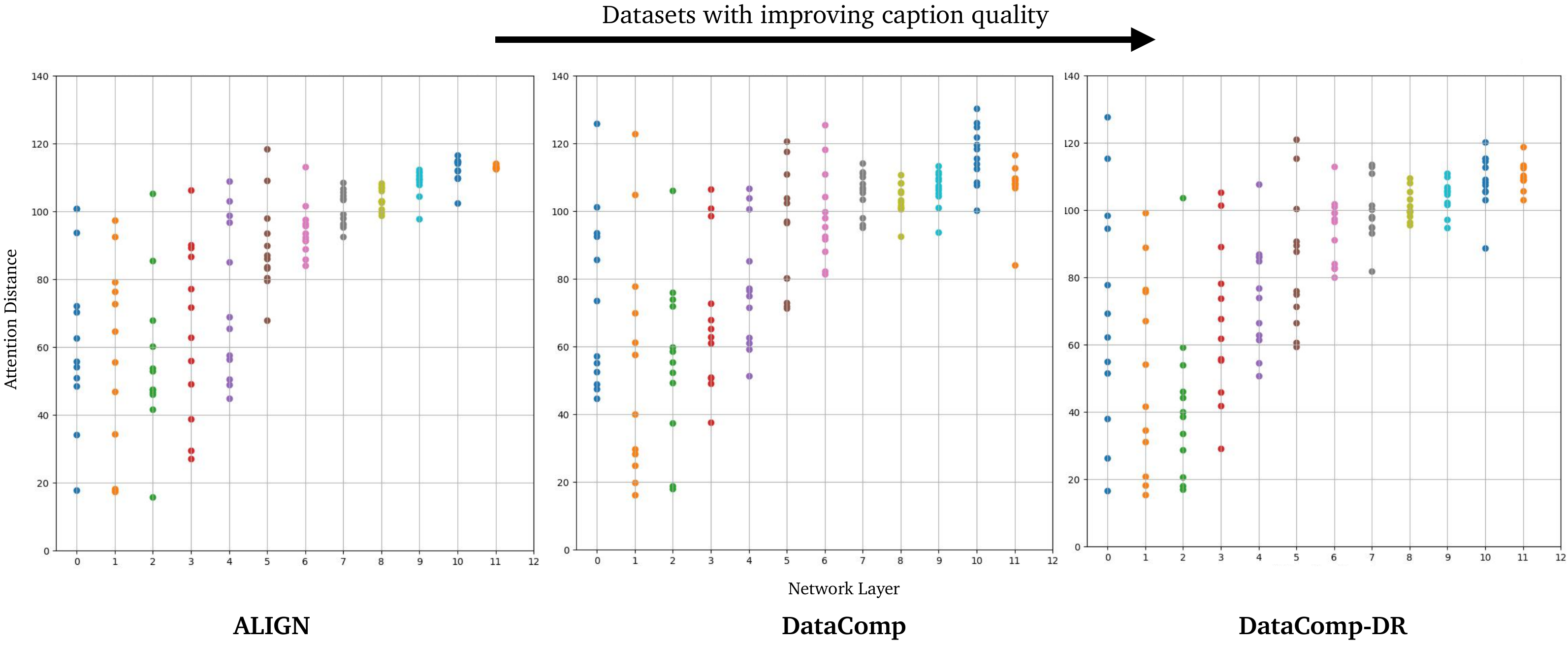}
    \caption{Average attention distances of ViT-B/16 model trained on 3 different datasets with varying caption quality. The caption quality improves from left to right. There is noticable improvement in diversity of attention distances when CLIP models are trained on datasets with better captions.}
    \label{fig:attn_distances}
\end{figure}

To understand the effects of quality of captions, we plot the average attention distance~\cite{vit} over the entire ImageNet-1K validation set. This helps us understand the information flow as it partially reflects the receptive field size of each attention head. The \texttt{[CLS]} token is ignored in the average computation as described in ~\cite{clipft}. From \cref{fig:attn_distances}, we notice a difference in average attention distance per head, especially for models trained on cleaner and more aligned captions in \datacomp{} and \datacompdr{}. For CLIP model trained on ALIGN dataset with noisy captions, we notice that there is less diversity in attention distances, especially towards the deeper layers. Intutively this indicates that there might be redundancies in the last layers and the model's capacity may not be fully utilized as described in~\cite{xie2023darkmim}. When better aligned captions are introduced in \datacompdr{}, we notice that attention heads tend to be more local while keeping the diversity within a layer. Local attention is more favorable for dense prediction tasks as observed in~\cite{clipft, dBOT}. Hence, CLIP models trained on captions that are less noisy and more aligned with the image tend to perform better on dense prediction tasks.  

\vspace{-5pt}
\section{Experiments}\label{sec:experiments}
\vspace{-5pt}

We evaluate the performance of CLIP model's vision encoder for 4 downstream tasks: Image Classification, Instance Segmentation, Semantic Segmentation and Depth Estimation. We primarily focus on end-to-end fine tuning performance for all the tasks. 

\noindent
\textbf{CLIP Pretraining}. We follow~\cite{mobileclip2024, lai2024veclip} in training CLIP models on \datacomp{} and \datacompdr{}. We minimize CLIP's contrastive loss on both ground-truth and synthetic captions as commonly done in~\cite{mobileclip2024, lai2024veclip}. We chose \datacomp{} and \datacompdr{} for experimentation due to its scale, i.e 1.28 billion image-text pairs. Other publicly available datasets like~\cite{lai2024veclip, fan2023laclip} only contain 200 to 400 million image-text pairs. Detailed list of hyperparameters are provided in the supplementary materials. For mobile architectures, we simply use the CLIP pretrained models on \datacompdr{} from~\cite{mobileclip2024}. 

\noindent
\textbf{Image Classification}. We fine-tune the vision encoder on ImageNet-1K dataset for 100 epochs following the settings in prior works like~\cite{clipft, dBOT}. More details on the exact setup will be provided in the supplementary materials. In \cref{tab:imagenet_results}, we compare supervised baselines (in gray) with recent self supervised and weakly supervised methods. 
From \cref{tab:imagenet_results}, it is evident that CLIP pre-training on datasets with large-scale good quality captions outperform recent state-of-the-art pretraining methods and other CLIP models trained on larger datasets with noisy captions.

% \vspace{15pt}
\noindent
\textbf{Object Detection and Instance Segmentation}. We report performance of all models with MaskRCNN~\cite{maskrcnn} head and Cascade-MaskRCNN~\cite{cascadercnn} head for instance segmentation on MS-COCO~\cite{MSCOCO} dataset. Models were trained using MMDetection library~\cite{mmdetection}. MaskRCNN models are trained using 1$\times$ schedule with single scale testing as described in~\cite{clipft}. Cascade-MaskRCNN models are trained using 3$\times$ schedule with single scale testing. We follow finetuning setup described in~\cite{dBOT, clipft}, more details in supplementary materials.
From~\Cref{tab:maskrcnn_results} and ~\cref{tab:cascade_maskrcnn_results}, ViT B/16 with CLIP pretraining on \datacomp{} and \datacompdr{} outperforms recent state-of-the-art pretraining methods.

% Full ImageNet results
\begin{table}[htb!]
    \centering
    \resizebox{0.65\linewidth}{!}{
        \begin{tabular}{l@{\hspace*{4mm}}c@{\hspace*{4mm}}c@{\hspace*{4mm}}c@{\hspace*{4mm}}c}
            \toprule[1.5pt]
            \multirow{2}{*}{\textbf{Pre-Training Method}} &
            \multirow{2}{*}{\textbf{Dataset}} &
            \textbf{Input Res.} &
            \textbf{Top-1 Acc.}
            \\
            & & \textbf{(px)} & \textbf{(\%)} \\
             \midrule[1.25pt]

            \textit{\textcolor{gray}{Supervised}} & & \\
            
            \textcolor{gray}{Baseline} & \textcolor{gray}{IN-21K} & 224 & \textcolor{gray}{84.0} \\
            Billion-Scale PT~\cite{billionscale}  & ANN-1.3B & 224 & 83.6 \\
            MOFI~\cite{mofi2024}              & I2E-1.1B & 224 & 84.1 \\   % BT: ewxaf57txx
            CatLIP~\cite{mehta2024catlip}                    & \datacomp{} & 224 & 84.3 \\
            CatLIP~\cite{mehta2024catlip}                    & \datacomp{} & 512 & 86.1 \\
            Scale-ViT~\cite{scalevit}         & JFT-3B  & 224 & \textbf{86.6} \\
            \midrule
            \textit{\textcolor{gray}{Self-Supervised Methods}} & & & \\
            BEiT~\cite{beit} & IN-22K & 224 & 83.2 \\
            MAE~\cite{mae}  & IN-1K & 224 & 83.6 \\
            MAE~\cite{singh2023effectiveness}  & IG-3B & 224 & 83.5 \\
            iBOT~\cite{ibot} & IN-1K & 224 & 84.0 \\
            dBOT~\cite{dBOT} & IN-1K & 224 & 84.5 \\
            MAWS~\cite{singh2023effectiveness} & IG-3B & 518 & \underline{86.4} \\
            \textit{\textcolor{gray}{Weakly Supervised Methods}} & & & \\
            FD-CLIP~\cite{clipft} & OpenAI-WIT + IN-1K & 224 & 85.0 \\
            SWAG~\cite{swag} & IG-3.6B & 384 & 85.3 \\
            \midrule[0.5pt]
            CLIP~\cite{clip} & OpenAI-WIT & 224 & 82.9 \\
            CLIP & ALIGN$^{\dagger}$ & 224 & 85.4 \\
            OpenCLIP~\cite{cherti2022reproducible} & LAION-2B & 224 & 85.5 \\
            CLIP-FT~\cite{dong2022ftclip} & OpenAI-WIT & 224 & 85.7 \\
            CLIP & \datacomp{}~\cite{datacomp} & 224 & 85.9 \\
            CLIP & \datacompdr{}~\cite{mobileclip2024} & 224 & \underline{86.4} \\
            CLIP & \datacompdr{}~\cite{mobileclip2024} & 384 & \textbf{87.2} \\
            \bottomrule[1.5pt]
        \end{tabular}
    }
    \caption{Results on ImageNet-1K validation set from end to end finetuning. All results are for ViT-B/16 architecture.$^{\dagger}$ Our reproduction of ALIGN~\cite{align} dataset.}
    \label{tab:imagenet_results}
\end{table}

\vspace{-5pt}
% Detection results
\begin{minipage}[t!]{0.47\textwidth}%
        \centering
        \resizebox{0.9\linewidth}{!}{
            \begin{tabular}{l@{\hspace*{5mm}} c@{\hspace*{5mm}}c@{\hspace*{5mm}}c}
                \toprule[1.5pt]
                \textbf{Method} & \textbf{Dataset} & \textbf{mAP}$^{box}$ & \textbf{mAP}$^{mask}$ \\
                \midrule
                % FD-DINO         & IN-1K & 46.1 & 40.9 \\
                CatLIP~\cite{mehta2024catlip} & \datacomp{} & 45.7 & 40.6 \\
                % FD-DeiT~\cite{clipft}         & IN-1K & 46.4 & 41.0 \\ 
                MAE~\cite{mae}  & IN-1K & 46.5 & 40.9 \\
                MAE~\cite{singh2023effectiveness}  & IG-3B & 46.4 & 42.1 \\
                MAWS~\cite{singh2023effectiveness} & IG-3B & 48.0 & 43.4 \\
                FD-CLIP         & OpenAI-WIT + IN-1K & 48.2 & 42.5 \\
                \midrule[0.5pt]
                CLIP            & OpenAI-WIT & 45.0 & 39.8 \\
                CLIP            & ALIGN      & 47.5 & 42.3 \\
                CLIP            & \datacomp{}      & 48.4 & 43.4 \\
                CLIP            & \datacompdr{} & \textbf{48.9} & \textbf{43.6} \\
               \bottomrule[1.5pt]
        \end{tabular}
        }
        \vspace{10px}
        \captionof{table}{Results on Mask-RCNN trained for 1$\times$ schedule. All results are for ViT-B/16 models.}\label{tab:maskrcnn_results}
\end{minipage}%
\hspace{5px}
\begin{minipage}[t!]{0.47\textwidth}%
        \centering
        \resizebox{0.8\linewidth}{!}{
            \begin{tabular}{l@{\hspace*{5mm}} c@{\hspace*{5mm}}c@{\hspace*{5mm}}c}
                \toprule[1.5pt]
                \textbf{Method} & \textbf{Dataset} & \textbf{mAP}$^{box}$ & \textbf{mAP}$^{mask}$ \\
                \midrule
                iBOT~\cite{ibot}    & IN-1K+IN-22K & 51.3 & 44.3 \\ 
                CatLIP~\cite{mehta2024catlip} & DataComp & 52.6 & 45.4 \\
                dBOT~\cite{dBOT}    & IN-1K & 52.7 & 45.7 \\  
                MAE~\cite{singh2023effectiveness}  & IG-3B & 53.8 & 46.5 \\
                MAWS~\cite{singh2023effectiveness} & IG-3B & 53.9 & 46.6 \\
                \midrule[0.5pt]
                CLIP    & OpenAI-WIT  & 49.7 & 43.0  \\
                CLIP    & ALIGN       & 53.5 & 46.1  \\
                CLIP    & \datacomp{}    & 53.9 & 46.4  \\
                CLIP    & \datacompdr{} & \textbf{54.6} & \textbf{47.1} \\
               \bottomrule[1.5pt]
        \end{tabular}
        }
        \vspace{10px}
        \captionof{table}{Results on Cascade Mask-RCNN trained for 3$\times$ schedule. All results are for ViT-B/16 models.}\label{tab:cascade_maskrcnn_results}
\end{minipage}%

% Segmentation and Depth estimation
\begin{minipage}[t!]{0.47\textwidth}%
        \centering
        \resizebox{0.9\linewidth}{!}{
            \begin{tabular}{l@{\hspace*{5mm}} c@{\hspace*{5mm}}c@{\hspace*{5mm}}c}
                \toprule[1.5pt]
                \textbf{Method} & \textbf{Dataset} & \textbf{mIoU} & \textbf{mAcc} \\
                \midrule
                % FD-DINO  & IN-1K              & 47.7 & - \\
                % FD-DeiT  & IN-1K              & 48.0 & - \\ 
                % MAE~\cite{singh2023effectiveness}  & IG-3B & 47.2 & 58.0 \\
                MAE~\cite{mae}                     & IN-1K & 48.1 & 58.9 \\
                iBOT~\cite{ibot} & IN-1K+IN-22K & 48.4 & 59.3 \\
                dBOT~\cite{dBOT} & IN-1K        & 49.5 & 60.7 \\
                MAWS~\cite{singh2023effectiveness} & IG-3B & 50.4 & 61.5 \\
                CatLIP~\cite{mehta2024catlip}   & \datacomp{} & 50.6 & 61.8 \\
                FD-CLIP~\cite{clipft}  & OpenAI-WIT + IN-1K & 51.7 & - \\
                \midrule[0.5pt]
                CLIP     & OpenAI-WIT         & 49.5 & - \\ 
                CLIP     & ALIGN              & 51.2 & 62.8 \\
                CLIP     & DFN+VeCLIP~\cite{lai2024veclip}         & 51.4 & 62.3 \\ 
                CLIP     & \datacomp{}           & 52.5 & 64.2 \\
                CLIP     & \datacompdr{}        & \textbf{53.9} & \textbf{65.2} \\
               \bottomrule[1.5pt]
        \end{tabular}
        }
        \vspace{10px}
        \captionof{table}{Semantic segmentation results on ADE20k using UperNet decoder. All results are for ViT-B/16 models.}\label{tab:segm_results}
\end{minipage}%
\hspace{5px}
\begin{minipage}[t!]{0.47\textwidth}%
        \centering
        \resizebox{0.8\linewidth}{!}{
            \begin{tabular}{l@{\hspace*{5mm}} c@{\hspace*{5mm}}c}
                \toprule[1.5pt]
                \textbf{Method} & \textbf{Dataset} & \textbf{RMSE}($\downarrow$) \\
                \midrule
                % FD-DeiT & IN-1K & 0.404 \\
                FD-DINO & IN-1K & 0.394 \\
                CatLIP~\cite{mehta2024catlip} & \datacomp{} & 0.394 \\
                MAE~\cite{mae}     & IN-1K & 0.383 \\
                MAWS~\cite{singh2023effectiveness} & IG-3B & 0.371 \\
                FD-CLIP~\cite{clipft} & OpenAI-WIT + IN-1K & 0.352 \\
                MAE~\cite{singh2023effectiveness}  & IG-3B & 0.348 \\
                \midrule[0.5pt]
                CLIP    & OpenAI-WIT  & 0.416 \\
                CLIP    & ALIGN & 0.380 \\
                CLIP    & \datacomp{} & 0.358  \\
                CLIP    & \datacompdr{} & \textbf{0.339} \\
               \bottomrule[1.5pt]
        \end{tabular}
        }
        \vspace{10px}
        \captionof{table}{Results on NYUv2 for depth estimation following the same settings as~\cite{xie2023darkmim}. All results are for ViT-B/16 models.}\label{tab:depth_results}
\end{minipage}%

\vspace{8pt}
\noindent
\textbf{Semantic Segmentation}. We use UperNet~\cite{upernet} head and follow the setup described in~\cite{dBOT}. Models were trained using MMSegmentation library~\cite{mmseg2020}. 
For mobile models, we use the SemanticFPN~\cite{kirillov2019panoptic} head and train the models using the setup described in~\cite{fastvit}. More details are provided in supplementary materials.
From ~\Cref{tab:segm_results}, CLIP pretraining significantly benefits segmentation task. In fact, CLIP trained on noisy ALIGN~\cite{align} dataset outperforms MAWS~\cite{singh2023effectiveness}. When pretrained on \datacompdr{}, we observe a significant $3.5$ (6.9\%) improvement in mIoU over MAWS~\cite{singh2023effectiveness}. 

\vspace{8pt}
\noindent
\textbf{Depth Estimation}. We report Root Mean Square Error (RMSE) on NYUv2 dataset~\cite{nyuv2}.
We use the same settings as described in ~\cite{clipft,xie2023darkmim}, more details are provided in supplementary materials.. From~\cref{tab:depth_results}, ViT-B/16 pretrained on \datacompdr{} outperforms recent state-of-the-art pretraining and multistage pretraining methods like~\cite{singh2023effectiveness, clipft}.

\begin{table}[t]
    \centering
        \resizebox{0.8\linewidth}{!}{
            \begin{tabular}{l@{\hspace*{5mm}} c@{\hspace*{5mm}}c@{\hspace*{5mm}}c@{\hspace*{5mm}}c@{\hspace*{5mm}}c}
                \toprule[1.5pt]
                \textbf{Encoder} & \textbf{Decoder} & \textbf{Pre-Training} & \textbf{Resolution} & \textbf{\# Params}(M) & \textbf{mIoU} \\
                \midrule

                \textcolor{gray}{InternImage-B}~\cite{internimage} & \textcolor{gray}{UperNet}~\cite{upernet} & \textcolor{gray}{Sup. IN-1K} &  \textcolor{gray}{512$\times$512} & \textcolor{gray}{128.0} & \textcolor{gray}{50.8} \\ 
                \textcolor{gray}{ViT-Adapter-B}~\cite{vitadapter} & \textcolor{gray}{SemanticFPN}~\cite{kirillov2019panoptic} & \textcolor{gray}{Sup. IN-22K} &  \textcolor{gray}{512$\times$512} & \textcolor{gray}{104.6} & \textcolor{gray}{50.7} \\
                \textcolor{gray}{ViT-Adapter-B}~\cite{vitadapter} & \textcolor{gray}{UperNet}~\cite{upernet} & \textcolor{gray}{Sup. IN-22K} &  \textcolor{gray}{512$\times$512} & \textcolor{gray}{133.9} & \textcolor{gray}{51.9} \\
                \textcolor{gray}{Swin-L}~\cite{swin} & \textcolor{gray}{UperNet}~\cite{upernet} & \textcolor{gray}{Sup. IN-22K} &  \textcolor{gray}{640$\times$640} & \textcolor{gray}{234.1} & \textcolor{gray}{52.1} \\

                \midrule[0.5pt]

                MCi0~\cite{mobileclip2024} & SemanticFPN~\cite{kirillov2019panoptic} & Sup. IN-1K & 512$\times$512 & 14.5 & 44.8 \\
                MCi1~\cite{mobileclip2024} & SemanticFPN~\cite{kirillov2019panoptic} & Sup. IN-1K & 512$\times$512 & 24.6 & 47.7 \\
                MCi2~\cite{mobileclip2024} & SemanticFPN~\cite{kirillov2019panoptic} & Sup. IN-1K & 512$\times$512 & 38.5 & 48.9 \\

                \midrule[0.5pt]
                
                MCi0~\cite{mobileclip2024} & SemanticFPN~\cite{kirillov2019panoptic} & CLIP \datacompdr{} & 512$\times$512 & 14.5 & 48.1 \textcolor{darkgreen}{(+3.3)} \\
                MCi1~\cite{mobileclip2024} & SemanticFPN~\cite{kirillov2019panoptic} & CLIP \datacompdr{} & 512$\times$512 & 24.6 & 49.8 \textcolor{darkgreen}{(+2.1)} \\
                MCi2~\cite{mobileclip2024} & SemanticFPN~\cite{kirillov2019panoptic} & CLIP \datacompdr{} & 512$\times$512 & 38.5 & 52.2 \textcolor{darkgreen}{(+3.3)} \\
                \bottomrule[1.5pt]
        \end{tabular}
        }
        \caption{Comparison pretraining methods for semantic segmentation on ADE-20k. For reference, we have included recent state-of-the-art semantic segmentation models (in gray). 
        }\label{tab:mobile_segm_results}
\end{table}

\begin{table}[t]
    \centering
        \resizebox{0.7\linewidth}{!}{
            \begin{tabular}{l@{\hspace*{5mm}}c@{\hspace*{5mm}}c@{\hspace*{5mm}}c@{\hspace*{5mm}}c}
                \toprule[1.5pt]
                \textbf{Model} & \textbf{Pre-Training} & \textbf{\# Params}(M) & \textbf{mAP}$^{box}$ & \textbf{mAP}$^{mask}$ \\
                \midrule
                
                \textcolor{gray}{ViT-Adapter-B}~\cite{vitadapter} & \textcolor{gray}{Sup. IN-1K} & \textcolor{gray}{284} & \textcolor{gray}{47.0} & \textcolor{gray}{41.8} \\
                \textcolor{gray}{InternImage-B}~\cite{internimage} & \textcolor{gray}{Sup. IN-1K} & \textcolor{gray}{115} & \textcolor{gray}{48.8} & \textcolor{gray}{44.0} \\
                \textcolor{gray}{ViT-Adapter-L}~\cite{vitadapter} & \textcolor{gray}{Sup. IN-22K} & \textcolor{gray}{347.9} & \textcolor{gray}{48.7} & \textcolor{gray}{43.3} \\
                
                \midrule[0.5pt]
                MCi0~\cite{mobileclip2024} & Sup. IN-1K & 31.0 & 41.8 & 38.0 \\
                MCi1~\cite{mobileclip2024} & Sup. IN-1K & 41.1 & 45.2 & 40.6 \\
                MCi2~\cite{mobileclip2024} & Sup. IN-1K & 55.0 & 46.6 & 41.7 \\
              
                \midrule[0.5pt]
                MCi0~\cite{mobileclip2024} & CLIP \datacompdr{} & 31.0 & 44.2 \textcolor{darkgreen}{(+2.4)} & 39.5 \textcolor{darkgreen}{(+1.5)} \\
                MCi1~\cite{mobileclip2024} & CLIP \datacompdr{} & 41.1 & 47.7 \textcolor{darkgreen}{(+2.5)} & 42.5 \textcolor{darkgreen}{(+1.9)} \\
                MCi2~\cite{mobileclip2024} & CLIP \datacompdr{} & 55.0 & 49.5 \textcolor{darkgreen}{(+2.9)} & 43.6 \textcolor{darkgreen}{(+1.9)} \\
             
                \bottomrule[1.5pt]
        \end{tabular}
        }
        \caption{Comparison pretraining methods for object detection task on MS-COCO using MaskRCNN~\cite{maskrcnn} detection head. All models are trained for 1$\times$ schedule. For reference we have included recent state-of-the-art object detection models (in gray). 
        }\label{tab:mobile_det_results}
\end{table}

% \vspace{-8pt}
\section{Impact of CLIP pretraining on Mobile Architectures}\label{sec:impact}
In the previous sections we have analyzed the impact of CLIP pretraining on \datacompdr{} for a larger architecture i.e. ViT-B/16. In this section, we analyze the benefits of CLIP pretraining on mobile architectures. Specifically, we finetune recently open-sourced MobileCLIP's image encoders~\cite{mobileclip2024}. We compare the widely used supervised pretraining on ImageNet-1K with CLIP pretraining on \datacompdr{}. In \cref{tab:mobile_segm_results}, we observe $3.3$ improvement in mIoU for the smallest architecture by CLIP pre-training on \datacompdr{}. MCi2 model obtains 52.2 mIoU, similar to much larger architectures like Swin-L and ViT-Adapter-B~\cite{vitadapter} models. In \cref{tab:mobile_det_results}, we observe $2.4$ improvement in bounding box mAP and $1.5$  improvement in mask mAP for the smallest model. MCi2 model obtains a bounding box mAP of 49.5, which is on par with larger architectures like ViT-Adapter-L~\cite{vitadapter} and performs even better than newer specialized architectures like InternImage-B~\cite{internimage}. 

% \vspace{-8pt}
\section{Conclusion}
In this work, we analyze the performance of CLIP pretraining for downstream dense prediction tasks. We find that scale of pretraining dataset and quality of captions make a significant difference. We systematically compare with MAE, a popular self-supervised pretraining method for vision transformers. In contrast to prior works, we showed that CLIP pretraining is highly competitive with MAE and even improves over MAE and recent MAWS pretraining methods at scale. We showed that large-scale CLIP pretraining on a dataset with good quality captions results in an image encoder that learns highly transferable representations. We also showed that CLIP pretraining benefits even smaller architectures.

\bibliography{egbib}

% % This will be its own pdf
\clearpage

\section*{Supplementary Materials - CLIP with Quality Captions: A Strong
Pretraining for Vision Tasks}

\section{Experimental Setup}\label{sec:exp_setup}
We evaluate the downstream task performance on four common benchmarks, image classification on ImageNet-1K dataset, object detection on MS COCO~\cite{MSCOCO}, semantic segmentation on ADE20k~\cite{ADE20K} and depth estimation on NYUv2~\cite{nyuv2}. All the downstream models were trained on a single node containing 8 NVIDIA A100 GPUs. In the following subsections we elaborate the various experimental setups.

\subsection{CLIP Pretraining}\label{sec:clip_pt_setup}
We perform large scale CLIP pretraining on our version of ALIGN, \datacomp{} and \datacompdr{} datasets using the hyperparameters listed in~\Cref{tab:hyperparams_clip_pt}. When training on \datacompdr{}, we use contrastive losses on both synthetic and real captions as done in~\cite{mobileclip2024,lai2024veclip}. We leverage the mobile architectures from~\cite{mobileclip2024}. These models are also trained on \datacompdr{} with additional distillation supervision. For further details on pre-training mobile architectures, please refer~\cite{mobileclip2024}.

\begin{table}[!htb]
\centering
\scalebox{0.85}{
\begin{tabular}{l|c}
\toprule
Hyperparameter   & Value \\
\midrule
Input resolution & 224$^2$\\
Context length & 77\\
Random resize crop scale & [0.9, 1.0] \\
Random resized crop ratio & [0.75, 1.33] \\
RangeAugment target value & (40, 20) \\
Train iterations & 200k \\
Warmup iterations & 2k \\
Global batch size & 65536 \\
Optimizer & AdamW \\
AdamW beta1 & 0.9\\
AdamW beta2 & 0.95\\
Max learning rate & 1e-3 \\
Min learning rate & 1e-6 \\
LR. decay schedule & cosine \\
Weight decay rate & 0.2 \\
Mixed precision& BFloat16 \\
EMA decay rate & 0.9995 \\
\bottomrule

\end{tabular}
}
\caption{Training hyperparameters for large-scale CLIP training.}
\label{tab:hyperparams_clip_pt}
\end{table}

\begin{table}[!htb]
\centering
\scalebox{0.85}{
\begin{tabular}{l|c c}
\toprule
\multirow{2}{*}{Hyperparameter}   & \multicolumn{2}{c}{Value} \\
& FT & LP \\
\midrule
Input resolution & \multicolumn{2}{c}{224$\times$224} \\
Data augmentation & \multicolumn{2}{c}{RRC} \\
Mixup $\alpha$ & \multicolumn{2}{c}{0.0}  \\
CutMix $\alpha$ & \multicolumn{2}{c}{0.0} \\
Random erase prob. & \multicolumn{2}{c}{0.0} \\
Label smoothing & \multicolumn{2}{c}{0.1} \\
Stochastic depth rate & \multicolumn{2}{c}{0.1}  \\
Train epochs & \multicolumn{2}{c}{100} \\
Warmup epochs & \multicolumn{2}{c}{5} \\
Batch size & \multicolumn{2}{c}{1024} \\
Optimizer & \multicolumn{2}{c}{AdamW} \\
Peak learning rate & 1e-4 & 5e-4\\
LR. decay schedule & \multicolumn{2}{c}{cosine} \\
Weight decay rate & 0.05 & 0.0005\\
EMA decay rate & \multicolumn{2}{c}{0.9995} \\

\bottomrule

\end{tabular}
}
\caption{Training hyperparameters for ImageNet-1k experiments. RRC $\rightarrow$ \texttt{RandomResizedCrop}, FT $\rightarrow$ full Fine-Tuning, LP $\rightarrow$ Linear Probing.}
\label{tab:hyperparams_inet1k}
\end{table}

\subsection{Image Classification}\label{sec:in1k_cls}
We fine-tune the vision encoder on ImageNet-1K dataset for 100 epochs following the settings in prior works like~\cite{clipft, dBOT}. Details of all the hyperparameters are listed in~\ref{tab:hyperparams_inet1k}. For linear probing, the encoder is frozen, we replace the CLIP projection layer with a learnable linear layer and train it using the setup described in~\ref{tab:hyperparams_inet1k}.

\subsection{Object Detection}\label{sec:obj_det}
In order to compare with existing literature, we train object detection models with both MaskRCNN and Cascade MaskRCNN detection heads. Along with detection, these models also perform instance segmentation. We follow the settings prescribed in recent works like~\cite{dBOT, clipft, singh2023effectiveness}. All evaluations reported in the main paper are from single-scale evaluations on MS COCO validation set following prior works. We sweep through stochastic depth rate in steps of \texttt{0.05} and peak learning rate for all the results reported in the paper and the ranges are listed in~\Cref{tab:hyperparams_det_coco}. For ViT-B/16 models, we use ViTDet style feature pyramid network. For mobile models from~\cite{mobileclip2024}, we follow the setup described in ~\cite{fastvit, efficientformer}. All models were trained using MMDetection library~\cite{mmdetection}.

\begin{table}[!htb]
\centering
\scalebox{0.85}{
\begin{tabular}{l|c c}
\toprule
 Hyperparameters & MaskRCNN & Cascade MaskRCNN \\
\midrule
Stochastic depth rate & \multicolumn{2}{c}{[0.0, ..., 0.3]}  \\
Data augmentation & \multicolumn{2}{c}{Multi scale RRC}  \\
Train epochs & 12 & 36 \\
Batch size & \multicolumn{2}{c}{16} \\
Optimizer & \multicolumn{2}{c}{AdamW} \\
Peak learning rate & \multicolumn{2}{c}{[5e-4, 2e-4, 1e-4]} \\
LR. decay schedule type & \multicolumn{2}{c}{Step-wise} \\
LR. decay schedule & [8, 11] & [27, 33] \\
Weight decay rate & 0.1 & 0.05 \\

\bottomrule

\end{tabular}
}
\caption{Training hyperparameters for object detection and instance segmentation experiments on MS COCO. RRC $\rightarrow$ \texttt{RandomResizedCrop}.}
\label{tab:hyperparams_det_coco}
\end{table}

\begin{table}[!htb]
\centering
\scalebox{0.85}{
\begin{tabular}{l|c c}
\toprule
 Hyperparameters & UperNet & SemanticFPN \\
\midrule
Stochastic depth rate & \multicolumn{2}{c}{[0.0, ..., 0.2]}  \\
Data augmentation & \multicolumn{2}{c}{RRC}  \\
Crop Size & \multicolumn{2}{c}{512$\times$512}  \\
Train iterations & 160k & 40k \\
Batch size & 16 & 64 \\
Optimizer & \multicolumn{2}{c}{AdamW} \\
Peak learning rate & \multicolumn{2}{c}{[5e-4, 2e-4, 1e-4]} \\
LR. decay schedule type & \multicolumn{2}{c}{Polynomial} \\
Warmup iterations & 1500 & - \\
Weight decay rate & 0.01 & 5e-4 \\

\bottomrule

\end{tabular}
}
\caption{Training hyperparameters for semantic segmentation experiments on ADE20k. RRC $\rightarrow$ \texttt{RandomResizedCrop}.}
\label{tab:hyperparams_segm}
\end{table}

\subsection{Semantic Segmentation}\label{sec:segm}
In order to compare with existing literature, we train segmentation models with UperNet and SemanticFPN heads. These models are trained on ADE20k~\cite{ADE20K} dataset following the settings prescribed in~\cite{dBOT, clipft, singh2023effectiveness}. All evaluations reported in the main paper are from single-scale evaluations on validation set following prior works. For ViT-B/16 models, we use ViTDet style feature pyramid network with UperNet head. For mobile models from~\cite{mobileclip2024}, we follow the setup described in ~\cite{fastvit, efficientformer} and train models with only SemanticFPN head. We sweep through stochastic depth rate in steps of \texttt{0.05} and peak learning rate for all the results reported in the paper and the ranges are listed in~\Cref{tab:hyperparams_segm}. All models were trained using MMSegmentation library~\cite{mmseg2020}.

\subsection{Depth Estimation}\label{sec:depth}
We follow the experimental setup and architecture as described in~\cite{xie2023darkmim, clipft}. The models are trained on NYUv2 dataset~\cite{nyuv2}. We sweep through stochastic depth rate in steps of \texttt{0.05} and peak learning rate for all the results reported in the paper and the ranges are listed in~\Cref{tab:hyperparams_depth}.

\begin{table}[!htb]
\centering
\scalebox{0.85}{
\begin{tabular}{l|c}
\toprule
 Hyperparameters & Value \\
\midrule
Stochastic depth rate & [0.0, ..., 0.2]  \\
Data augmentation &RRC  \\
Crop Size & 480$\times$480  \\
Train epochs & 25 \\
Batch size & 24 \\
Optimizer & AdamW \\
Peak learning rate & [7e-4, 5e-4, 2e-4, 1e-4] \\
Layer decay rate & 0.8 \\
Weight decay rate & 0.05 \\

\bottomrule

\end{tabular}
}
\caption{Training hyperparameters for depth estimation experiments on NYUv2 dataset. RRC $\rightarrow$ \texttt{RandomResizedCrop}.}
\label{tab:hyperparams_depth}
\end{table}

\end{document}

%% file: defs.tex
\newcommand{\comment}[1]{}

\def\ee{{\boldsymbol e}}

\def\vv{{\boldsymbol v}}

\def\UU{{\boldsymbol U}}

\definecolor{colorYes}{RGB}{51,160,44}
\definecolor{colorNo}{RGB}{228,26,28} %

\newcommand{\cmark}{\textcolor{colorYes}{\ding{51}}}%
\newcommand{\xmark}{\textcolor{colorNo}{\ding{55}}}%

\def\datacomp{{DataComp}}

\newcommand{\LCEIT}{{\mathcal{L}^{\text{I2T}}}}
\newcommand{\LCETI}{{\mathcal{L}^{\text{T2I}}}}

\newcommand{\Similarity}{{\mathcal{S}}}
\newcommand{\FeatImgi}{{\varPhi_{\text{img}}^{(i)}}}
\newcommand{\FeatImg}{{\varPhi_{\text{img}}}}
\newcommand{\FeatTxt}{{\varPhi_{\text{txt}}}}